\documentclass{article}

\PassOptionsToPackage{numbers, compress}{natbib}

\usepackage[final]{nips_2018}

\usepackage[utf8]{inputenc} 
\usepackage[T1]{fontenc}    
\usepackage{hyperref}       
\usepackage{url}            
\usepackage{booktabs}       
\usepackage{amsfonts}       
\usepackage{nicefrac}       
\usepackage{microtype}      
\usepackage{amsmath}
\usepackage{graphicx}
\usepackage{subcaption}
\usepackage{textcomp}
\usepackage[export]{adjustbox}

\setcitestyle{square}

\title{Slum Segmentation and Change Detection : A Deep Learning Approach}

\author{
    Shishira R Maiya\thanks{Joint First authors.} \\
    Robert Bosch Center for Cyber Physical Systems \\
    Indian Institute of Science\\
    Bangalore, Karnataka 560054 \\
   \texttt{shishirar@iisc.ac.in} \\
   \And
    Sudharshan Chandra Babu\footnotemark[1]\\
    Department of Computer Science\\
    Indian Institute of Technology Bombay\\
    Mumbai, Maharashtra 400076 \\
\texttt{cbsudu@gmail.com} \\
}

\begin{document}

\maketitle

\begin{abstract}
More than one billion people live in slums around the world. In some developing countries, slum residents make up for more than half of the population and lack reliable sanitation services, clean water, electricity, other basic services. Thus, slum rehabilitation and improvement is an important global challenge, and a significant amount of effort and resources have been put into this endeavor. These initiatives rely heavily on slum mapping and monitoring, and it is essential to have robust and efficient methods for mapping and monitoring existing slum settlements. In this work, we introduce an approach to segment and map individual slums from satellite imagery, leveraging regional convolutional neural networks for instance segmentation using transfer learning. In addition, we also introduce  a method to perform change detection and monitor slum change over time. We show that our approach effectively learns slum shape and appearance, and demonstrates strong quantitative results, resulting in a maximum AP of 80.0.

\end{abstract}

\section{Introduction}

Currently, about one-quarter of the world’s urban population live in slums \cite{unhabitat}. These slum residents lack basic resources such as clean water, proper sanitation, electricity, and other necessary basic services. Various initiatives have been undertaken by international organizations and world governments in the past few decades towards slum improvement and rehabilitation. These initiatives rely heavily on the information provided by slum mapping and monitoring, such as scale, boundaries and slum growth, crucial for slum policy planning and development. Thus it is essential to have automated, robust, and efficient methods for slum mapping and monitoring. Slums differ greatly in terms of shape and appearance. Current approaches to slum segmentation and change detection \citep{Kuffer2016ExtractionOS,Eckert2011UrbanEA,doi:10.1179/136821909X12581187860130} are limited and do not adapt very well to the variance in shape and texture. In addition, these approaches classify all detected instances of slums as one entity, given a satellite image, and do not recognize individual slums in a satellite image, which is essential for developing rehabilitation strategies for individual slums.

We concentrate our work on the slums in Mumbai--Dharavi, The Mankhurd-Govandi belt, Kurla-Ghatkopar belt, Dindoshi and The Bhandup-Mulund slums. The number of slum-dwellers in Mumbai is estimated to be around 9 million, up from 6 million in 2001 that is, 62\% of of Mumbai live in informal slums \citep{JainBhavika}.

In our work we introduce the following contributions:
\begin{enumerate}
  \item We propose an instance segmentation based approach to the problem of slum mapping, that recognizes each slum in a given image, leveraging transfer learning, without the need for a large dataset. Our approach is based on the Mask R-CNN \cite{He2017MaskR} framework, and automatically recognizes and segments individual slums from satellite imagery. In order to study this, we curate a custom dataset, consisting of satellite images of slums along with their polygon masks.
  \item We introduce a method to monitor slum size increase or decrease over time and perform change detection. 
\end{enumerate}

 We show that our method effectively identifies individual slums and demonstrates good qualitative and quantitative results.

\section{Related Work}

Early work on slum segmentation and mapping, include those based on  object-based image analysis (OBIA) and texture-based methods \citep{Kuffer2016SlumsFS,Kohli2013TransferabilityOO, doi:10.1080/01431161.2010.523727}. In texture-based methods, the co-occurrence matrix (GLCM) is commonly used \citep{Kuffer2016ExtractionOS, Eckert2011UrbanEA, doi:10.1179/136821909X12581187860130}. They're limited by the fact that the extraction of a specific feature depends on the technique used. In addition, they have parameters that need to be optimized through trial and error \citep{6297992,DellAcqua2006UnstructuredHS,7005456}. In addition, these approaches classify all the slums as one entity in a single satellite image, and do not identify individual instances of slums.

Deep convolutional networks for segmentation have demonstrated strong results over other approaches \cite{Lin2014MicrosoftCC}, and can work well in the problem of slum mapping. Change detection is the process of identifying differences by observing images at different times \citep{Bruzzone2013ANF, changedetection2, changedetectionmain}. Change detection on satellite imagery can provide important insights on urban development, and growth of informal and formal settlements \citep{changedetection2}.

Our approach identifies individual instances of slum in a given image using the Mask R-CNN model and we show that it has sufficient capacity to learn the visual and spatial features about slum settlements in satellite imagery. The model captures the inherent visual distribution sufficiently, and overcomes the above limitations of other approaches. Our approach to change detection is straightforward and builds upon other popular approaches \citep{changedetection2,changedetectionmain}.

\section{Approach}

\subsection{Dataset}

\begin{figure}[h!]
  \centering
  \begin{subfigure}[b]{0.48\linewidth}
    \includegraphics[width=\linewidth]{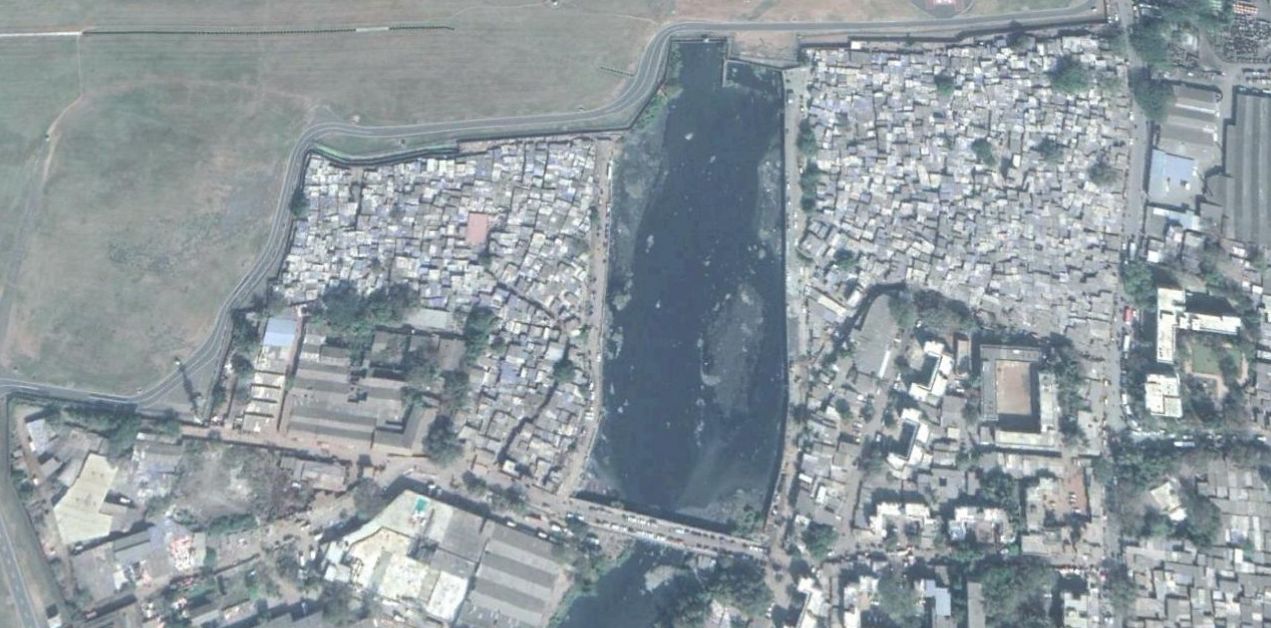}
  \end{subfigure}
  \begin{subfigure}[b]{0.48\linewidth}
    \includegraphics[width=\linewidth]{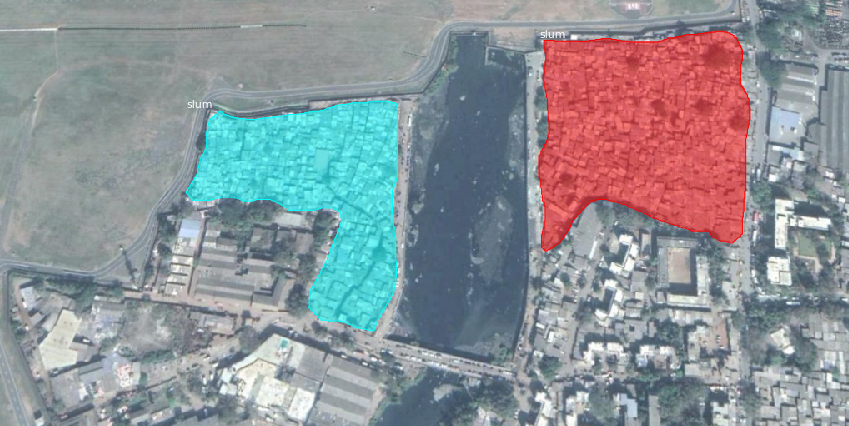}
  \end{subfigure}
  \caption{Example ground truth image (left) and segmentation mask (right) at a scale of 100 m. Map data
{\textcopyright}  2018 Google, DigitalGlobe.}
  \label{fig:dataset}
\end{figure}

We curated a dataset containing 3-band (RGB) satellite imagery with 65 cm per pixel resolution collected from Google Earth. Each image has a pixel size of 1280x720. The dataset consists of satellite images in two scales--100 m and 1000 m. These two scales provide different features that are used in small-scale and large-scale slum analysis respectively. The satellite imagery covers most of Mumbai and we include images from 2002 to 2018, to analyze slum change. Variability in resolutions of older images exist, due to the difference in satellites. Each image contains slum(s) along with formal settlements and vegetation. All images in the dataset have a paired list of polygons that describes sum instances. To verify our annotations, we used data provided by the Slum Rehabilitation Authority of Mumbai (SRA). We used 513 images for training, and 97 images for testing, for each scale. An example image at the 100 m scale and it's ground truth is depicted in Figure \ref{fig:dataset}.

\subsection{Slum Segmentation}

Our approach to slum segmentation is based on Mask R-CNN, a powerful and flexible instance segmentation model. The Mask R-CNN model consists of  two stages--the first stage scans the image and generates region based proposals, and the second stage classifies the proposals and generates bounding boxes and masks. The Mask R-CNN architecture we use is based on a ResNet-101 \citep{He2016DeepRL} and a Feature Pyramid Network \cite{Lin2017FeaturePN} backbone. We train the Mask R-CNN by optimizing the following multi-task loss
function, which combines the classification, localization and segmentation mask loss.

\begin{equation*}
\mathcal{L} = \mathcal{L}_\text{cls} + \mathcal{L}_\text{box} + \mathcal{L}_\text{mask}
\end{equation*}

where $\mathcal{L}_\text{cls}$ and $\mathcal{L}_\text{box}$ are the classification loss and bounding box regression loss respectively.The mask branch of the network generates a m x m dimensional mask for each Region of interest (RoI) and for each class,with K classes in total. Thus the resulting output tensor size is $K \cdot m^2$. $\mathcal{L}_\text{mask}$ is the average binary cross-entropy loss,
which includes the k-th mask in the region is mapped with the ground truth class k.

\begin{equation*}
\mathcal{L}_\text{mask} = - \frac{1}{m^2} \sum_{1 \leq i, j \leq m} \big[ y_{ij} \log \hat{y}^k_{ij} + (1-y_{ij}) \log (1- \hat{y}^k_{ij}) \big]
\end{equation*}

The input to the model is a satellite image, and the outputs are the bounding boxes, predicted masks and the confidence score. We train two models, one for each scale. We leverage transfer learning and pre-train the network on the COCO dataset \citep{Lin2014MicrosoftCC}.

\subsection{Slum change detection}

The input consists of a pair of satellite images, representing the same location, but at different points of time. For detecting change in the size of slums, we follow a two stage approach--We first pass both the images through the Mask R-CNN and predict masks for each image. We then subtract the binary masks and obtain a percentage increase or decrease.

\subsection{Training details}

We used an open source implementation of Mask R-CNN \cite{matterport_maskrcnn_2017}. We fine-tuned on the pre-trained Mask R-CNN network for 128 epochs, with a batch size of 2. We used the Adam optimizer \citep{DBLP:journals/corr/KingmaB14} with an initial learning rate of $10^{-4}$, and we decayed the learning rate at 50 and 120 epochs by a factor of 10. We padded and resized each image to 1024x1024, without changing the aspect ratio. We performed data augmentation by  horizontal and vertical flipping, rotation, translation, and variations in hue and saturation. We trained the model for around 4 hours on a Nvidia 1080 Ti GPU. We trained two models for each scale, with minor changes in the training process.
    
\section{Results}

\begin{figure}[h!]
  \centering
  \begin{subfigure}[b]{0.32\linewidth}
    \includegraphics[width=\linewidth]{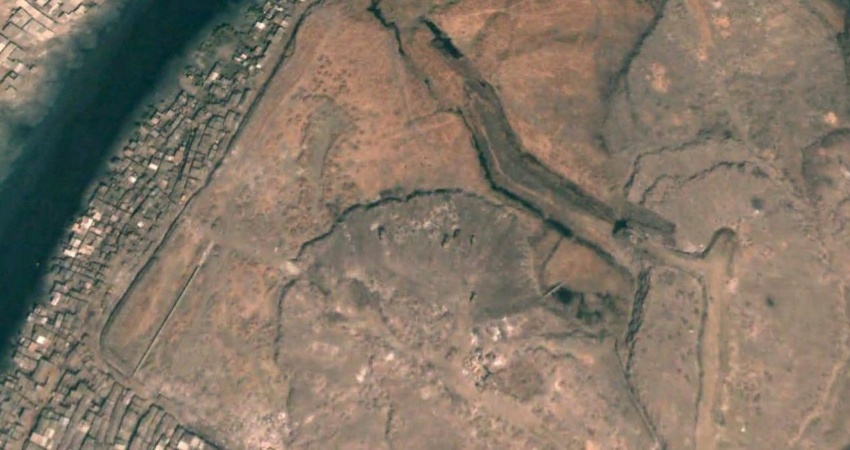}
  \end{subfigure}
  \begin{subfigure}[b]{0.32\linewidth}
    \includegraphics[width=\linewidth]{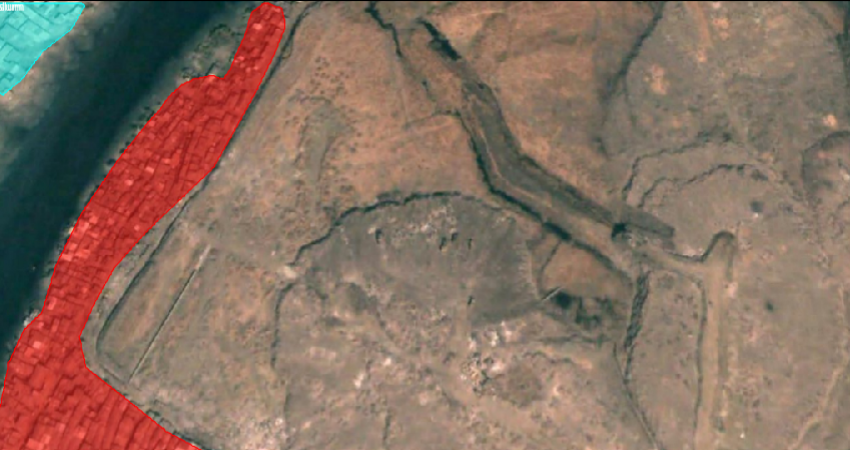}
  \end{subfigure}
  \begin{subfigure}[b]{0.32\linewidth}
    \includegraphics[width=\linewidth]{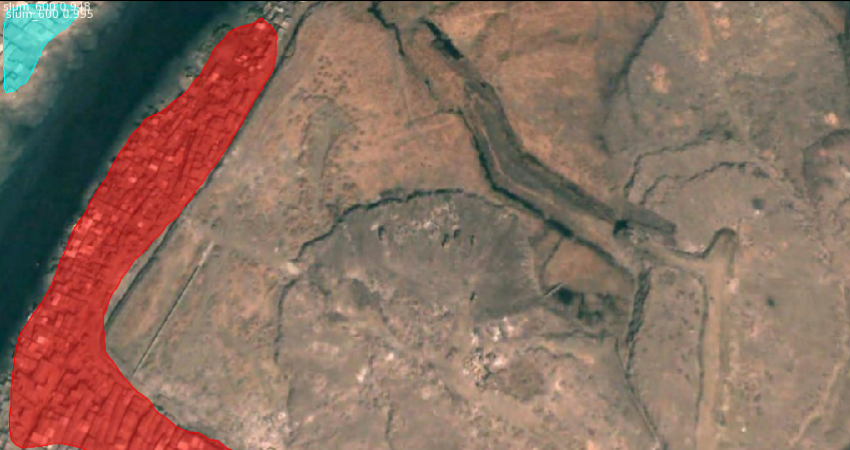}
  \end{subfigure}
    \begin{subfigure}[b]{0.32\linewidth}
    \includegraphics[width=\linewidth]{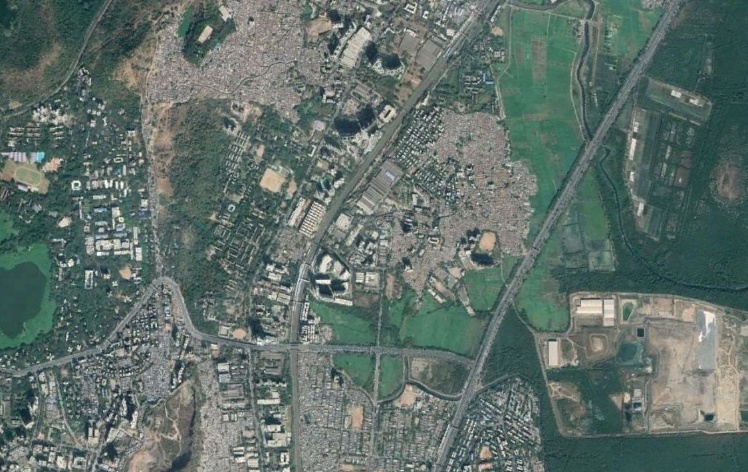}
  \end{subfigure}
  \begin{subfigure}[b]{0.32\linewidth}
    \includegraphics[width=\linewidth]{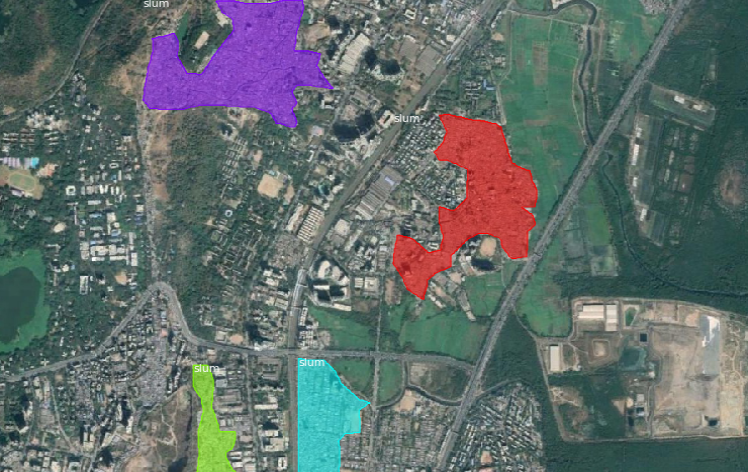}
  \end{subfigure}
  \begin{subfigure}[b]{0.32\linewidth}
    \includegraphics[width=\linewidth]{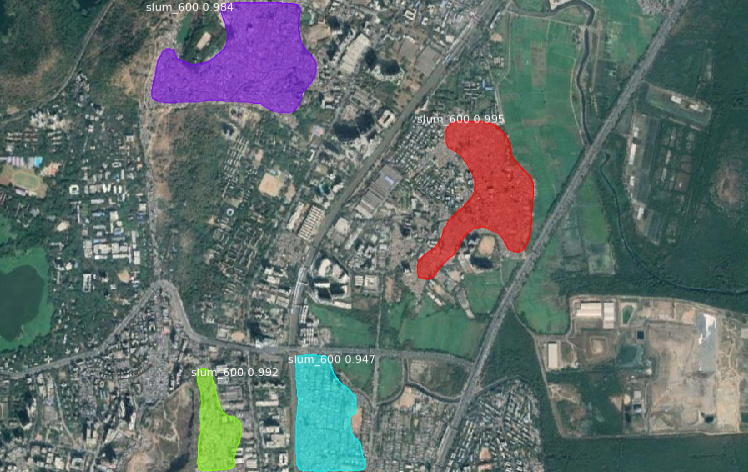} 
  \end{subfigure}
  \caption{ \textbf{Top:} Results on 100 m scale. \textbf{Bottom:}  Results on 1000 m scale. It can be observed that the 1000 m images have less learnable visual features. Both the rows contain input images from the test set (left), Ground truth masks (center), and predicted masks (right). Map Data {\textcopyright} 2018 Google, DigitalGlobe.}
  \label{fig:slumseg}
\end{figure}

\begin{figure}[h!]
  \centering
  \begin{subfigure}[b]{0.32\linewidth}
    \includegraphics[width=\linewidth]{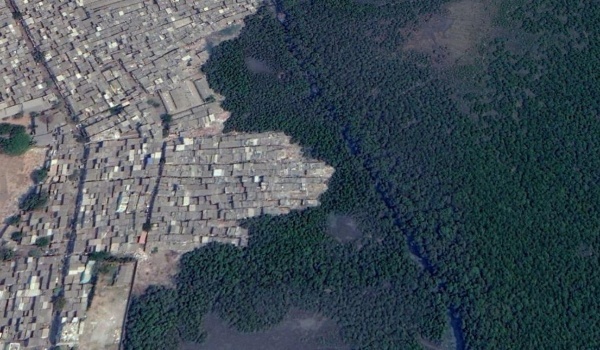}
  \end{subfigure}
  \begin{subfigure}[b]{0.32\linewidth}
    \includegraphics[width=\linewidth]{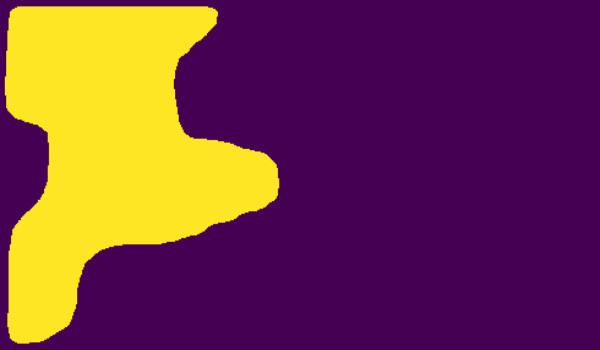}
  \end{subfigure}
  \begin{subfigure}[b]{0.32\linewidth}
    \includegraphics[width=\linewidth]{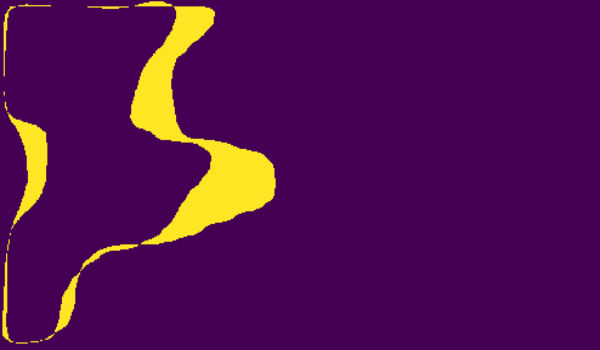}
  \end{subfigure}
    \begin{subfigure}[b]{0.32\linewidth}
    \raggedleft
    \includegraphics[width=\textwidth]{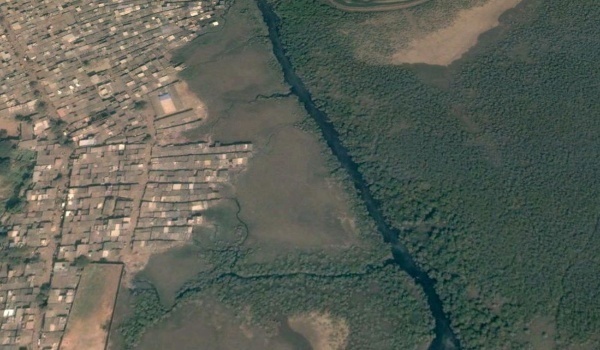}
  \end{subfigure}
  \begin{subfigure}[b]{0.32\linewidth}
    \includegraphics[width=\linewidth]{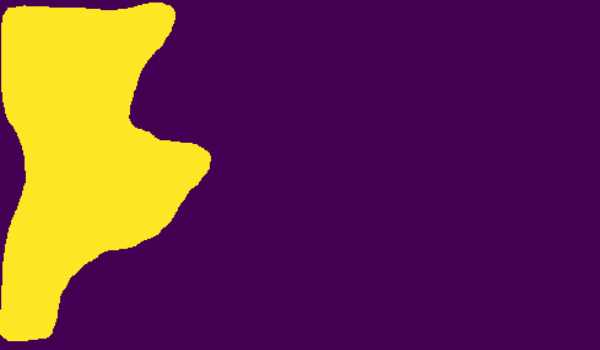}
  \end{subfigure}
  \begin{subfigure}[b]{0.32\linewidth}
    \includegraphics[width=\linewidth]{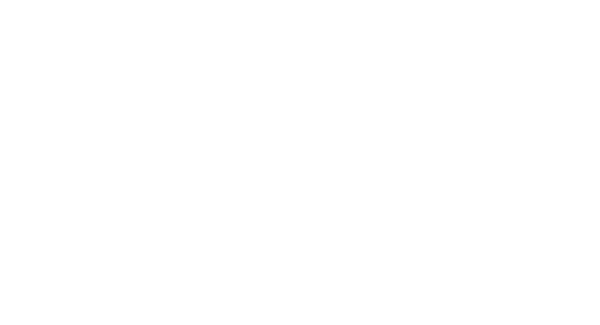}
  \end{subfigure}

  \caption{ 35.25\% change between 2018 (top left) and 2005 (bottom left). Predicted masks (middle) and change subtraction map (top right). Map Data {\textcopyright} 2004 Google, Copernicus} Map Data {\textcopyright} 2018 Google,DigitalGlobe. 
  \label{fig:changedetection}
\end{figure}

\subsection{Slum segmentation}

We evaluate our approach on two metrics -- IoU (Intersection over Union) and AP$_{\text{50}}$ (Average Precision at 50\% overlap), commonly used in segmentation. We demonstrate strong quantitative and qualitative results on these metrics on the test set, and recognize individual slum instances, as illustrated in Table 1 and in Figure \ref{fig:slumseg}. We find that the model performs better on the images from 100 m scale, due to the larger number of visual features (slum huts, boundaries etc) available in the image. We also discover that the model shows satisfactory results on certain slum regions. These instances are quite heterogeneous and contain multiple buildings and vegetation within the slum.

\subsection{Slum change detection}

We predict the percentage change of the slums in a given satellite image. Figure \ref{fig:changedetection} shows the change on a test image between 2005 and 2018. (35.25\% change)

 \begin{table}[h!]
 \caption{Results on test set and and slum-wise analysis}

 \begin{center}
 \begin{tabular}{l l l l l}
 \toprule
 &\multicolumn{2}{c}{100 m} & \multicolumn{2}{c}{1000 m} \\ 
\textbf{Slums} &\textbf{IoU}& \textbf{AP$_{\text{50}}$}    & \textbf{IoU} &\textbf{AP$_{\text{50}}$}      \\
     \midrule
      Test Set  & 0.86 & 80.2  & 0.73 & 38.3\\
      Govandi  & 0.89 & 60.3 & 0.80 & 59.2\\
      Bhandup  & 0.88 & 95.2 & 0.78 & 75.9 \\
      Dharavi  & 0.90 & 75.4 & 0.67 & 15.5  \\
      \bottomrule
\end{tabular}
\end{center}
\end{table}

\section{Conclusion}

In this work, we present an instance segmentation based approach to address the problem of slum mapping and monitoring. We show that our method achieves strong performance on these tasks. We hope that our work will be useful to organizations and other entities in their slum improvement and rehabilitation initiatives.

\bibliography{references}
\bibliographystyle{ieeetr}

\end{document}